\documentclass[letterpaper]{article}
\usepackage[preprint]{aaai2027}
\usepackage[hyphens]{url}
\usepackage{graphicx}
\usepackage{natbib}
\usepackage{caption}
\usepackage{newfloat}
\usepackage{listings}
\DeclareCaptionStyle{ruled}{labelfont=normalfont,labelsep=colon,strut=off}
\usepackage{booktabs}
\usepackage{amsmath}
\usepackage{tabularx}
\usepackage{tikz}
\usetikzlibrary{positioning,arrows.meta,calc}

\newcolumntype{L}{>{\raggedright\arraybackslash}X}
\newcolumntype{C}{>{\centering\arraybackslash}X}
\newcommand{\CE}{\mathrm{CE}}

\title{Same Place, Different Answer: Why LLM Estimates Don't Add Up across Spatial Scales}
\author{
    Devansh Shah\textsuperscript{\rm 1},
    Aviral Joshi\textsuperscript{\rm 2},
    Anirudh Joshi\textsuperscript{\rm 2}
}
\affiliations{
    \textsuperscript{\rm 1}University of Southern California\\
    \textsuperscript{\rm 2}Carnegie Mellon University\\
    devansh@alumni.usc.edu
}

\begin{document}

\maketitle

\begin{abstract}
Census geography supplies an exact symbolic constraint: the population of a tract equals the sum of the populations of its block groups. We use this constraint to test the consistency of large language model (LLM) estimates without ground truth. Queried one area at a time about 30 Cook County tracts and their 89 block groups, two frontier reasoning models returned block-group estimates whose sum, averaged over three runs, differed from their own tract estimates by a median of 12.0\% and 12.6\%, against 4.8--6.4\% attributable to run-to-run noise; all seven further settings also exceeded their noise floors. A pre-specified confirmatory study of 300 tracts and 964 block groups, scored against exact 2020 Census counts, replicated the violation in all three settings and showed that the same area receives estimates 17--29\% lower when presented as a block group rather than a tract. The census contains none of these effects: they are spurious scale effects produced by the models. Shrinking each area toward a typical size reproduces the violation, and its size tracks tract error across settings (Spearman $\rho$ 0.83). Enforcing the constraint after generation, by reconciling the two levels, lowered tract error in every setting, whereas requesting both levels in a single call satisfied the constraint, usually at a cost in block-group accuracy.
\end{abstract}

\section{Introduction}\label{sec:intro}

Large language models (LLMs) are increasingly used to supply quantities attached to places: the population of a neighbourhood, the income of a district, the number of housing units in a block. Such estimates can feed maps, rankings and screening tools, including for areas without published statistics~\cite{geollm,manvi2024bias}. Area statistics carry structure that a model must respect. Census geography is hierarchical: block groups partition census tracts, so the population of a tract equals the sum of the populations of its block groups. The constraint is exact, known in advance and independent of the true values.

This paper uses that constraint to evaluate LLM estimates without ground truth. A model asked about a tract and, in separate calls, about each of its block groups must return answers that add up; any discrepancy is the model's own error. The check is an instance of metamorphic testing~\cite{llmorph} and of consistency checking for models whose outputs are hard to verify directly~\cite{fluri2023,paleka2024}. Geographers know the underlying phenomenon as the modifiable areal unit problem (MAUP): statistics computed on areas change when the same data are aggregated at a different scale~\cite{gehlke1934,openshaw1979,openshaw1984}. For counts, the true scale effect between nested levels is zero by construction, so every scale effect in a model's answers is spurious. A recent review of governance risks in LLM-based GIS calls for ``MAUP/scale sensitivity checks''~\cite{governance}; we found no study that runs one on a language model.

The constraint also serves a neuro-symbolic purpose. It acts as a verifier, flagging incoherent outputs without labels, and as a repair operator, mapping incoherent outputs to coherent ones after generation, as hierarchical forecast reconciliation does for statistical forecasts~\cite{hyndman2011,wickramasuriya2019}. We compare this post-hoc enforcement with joint prompting, which shows the model all levels in a single call and leaves the constraint to it. Our contributions are:
\begin{enumerate}
  \item \textbf{A label-free test for nested area estimates}, based on the additive census constraint, with a noise model that separates model inconsistency from run-to-run variation and an identity test for areas that occur at two levels of the hierarchy (Section~\ref{sec:method}).
  \item \textbf{Evidence of spurious scale effects in all nine settings.} Across nine settings (models at given reasoning efforts) on 30 tracts, averaged block-group estimates miss the models' own tract estimates by more than run-to-run noise allows (Study~1). A pre-specified confirmatory study of 300 tracts, scored against exact counts, replicates the violation, shows that it grows with the number of block groups, and finds that the same area receives estimates 17--29\% lower when presented as a block group than as a tract (Study~2).
  \item \textbf{A mechanism}: shrinkage of each area toward a typical size for its level, which regressions given the same inputs reproduce (Section~\ref{sec:why}).
  \item \textbf{A comparison of repairs}: reconciliation after generation lowers tract error in every setting, while single-call prompting satisfies the constraint at a cost in block-group accuracy in most settings (Section~\ref{sec:repair}).
\end{enumerate}

\section{Related Work}\label{sec:related}

\paragraph{Geographic knowledge in LLMs.} Bhandari et al.\ probe what LLMs know about places, and Manvi et al.\ find their geographic estimates biased~\cite{bhandari2023,manvi2024bias}. GeoLLM showed that adding nearby OpenStreetMap places to the prompt markedly improves estimates of population and income~\cite{geollm}. GPSBench, GeoBenchX and GISAgentBench evaluate coordinate reasoning and GIS agents~\cite{gpsbench,geobenchx,gisagentbench}. A recent audit of LLM priors at block-group level uses a single unit throughout~\cite{neighborhood}, and Guo et al.\ report that GPT-5.4, prompted with mobility context, ``tends to over-smooth locally, assigning similar values to neighboring CBGs''~\cite{guo2026}. These evaluations score each task in a single form at a single spatial unit; none checks whether answers agree across levels.

\paragraph{Consistency and constraints.} Elazar et al.\ test whether language models give the same fact under paraphrase~\cite{elazar2021}, and LLMorph applies metamorphic testing to language tasks~\cite{llmorph,cho2025icsme}. Fluri et al.\ argue for consistency checks where no ground truth is available~\cite{fluri2023}, and Paleka et al.\ find that the consistency of LLM forecasters across logically related questions correlates with their accuracy~\cite{paleka2024}. Neuro-symbolic methods inject constraints into learning, for instance through a semantic loss~\cite{xu2018semantic} or through output layers that make hierarchical multi-label predictions coherent by construction~\cite{giunchiglia2020}. Closest to our repairs, BeliefBank and ConCoRD enforce logical consistency on a frozen model's outputs after generation with a MaxSAT solver, which improves accuracy~\cite{kassner2021beliefbank,mitchell2022concord}. Wolf et al.\ partition populations into verbalized subpopulations and find that LLM estimates violate the law of total probability; in persona prompting, estimates aggregated from finer subpopulations often match reference data better than direct population-level estimates, a pattern they call the macro fallacy~\cite{wolf2026partition}. Kotawala bounds the incoherence of probabilistic claims composed by multi-component LLM agents and repairs it by projection, whereas retrieval, partition-aware prompting and an aggregator LLM fail or regress~\cite{kotawala2026locally}. Census geography adds an exact physical partition with additive counts, reference values at every level, and areas that occur at two levels, which give an identity test. We study the numeric case: a linear equality constraint on a frozen model's estimates, enforced by reconciliation.

\paragraph{Coherent estimates for nested areas.} Making nested estimates agree is an established problem in statistics. Small-area estimation benchmarks model-based estimates so that they sum to a reliable total~\cite{wang2008sae,pfeffermann2013}; pycnophylactic interpolation preserves zone totals~\cite{tobler1979}; and hierarchical forecasting reconciles incoherent forecasts~\cite{hyndman2011,wickramasuriya2019}. The MAUP literature documents scale and zoning effects on census statistics~\cite{gehlke1934,openshaw1979,fotheringham1991}. We bring these tools to LLM estimates, where incoherence is a property of the estimator rather than of the data.

\section{Constraint, Measures and Repairs}\label{sec:method}

\paragraph{The additive constraint.} Let tract $p$ contain the block groups $C(p)$, with $k = |C(p)|$. Census counts satisfy
\begin{equation}
y_p = \sum_{c \in C(p)} y_c \quad \text{for every tract } p,
\label{eq:constraint}
\end{equation}
a linear system $A\mathbf{y} = \mathbf{0}$ over the vector of all area values. A model queried about each area in a separate call returns estimates $\hat{\mathbf{y}}$. With $S_p = \sum_{c \in C(p)} \hat y_c$ the sum of its block-group answers, its \textbf{coherence error} for tract $p$ is the relative violation of the constraint:
\begin{equation}
\CE(p) = \frac{\lvert S_p - \hat y_p \rvert}{\hat y_p}.
\label{eq:ce}
\end{equation}
The signed coherence error omits the absolute value. The coherence error requires no ground truth, so the check $\CE(p) > \tau$ acts as a label-free verifier; within settings we use $\tau = 10\%$. Because census counts satisfy \eqref{eq:constraint} exactly, any coherence error is a scale effect produced by the model, which we call a \textbf{spurious scale effect}. When $k = 1$, the tract and its block group are the same area and the constraint reduces to an identity: the signed coherence error then measures how much the answer depends on whether the area is presented as a tract or as a block group. A tract is scored only when its tract answer and all its block-group answers are present; zeros and refusals count as missing.

\paragraph{Separating inconsistency from noise.} Repeated identical queries return different numbers, and noisy answers violate the constraint even when the underlying estimates are coherent. To bound this effect, we make each model's averaged answers add up exactly, add back each area's own run-to-run deviations, average as the runs are averaged, and record the median coherence error in each of 1,000 simulations (Appendix~\ref{app:noise}). An observed coherence error beyond the upper tail of this \emph{noise-alone} distribution cannot be explained by run-to-run variation; Table~\ref{tab:s1settings} reports the distribution's median and Table~\ref{tab:s2} its 97.5th percentile.

\paragraph{Scale effects on conclusions.} Median income does not add up, so for income we compare a conclusion drawn at each level: the ten tracts a model ranks poorest from its tract answers, and the ten it ranks poorest from its block-group answers, averaged within each tract and weighted by the model's own block-group population estimates. Their \textbf{poorest-ten overlap} is the Jaccard index of the two lists; we also report the Spearman correlation between the two rankings. The same measures computed from census values give the true scale effect, and any difference from it is spurious.

\paragraph{Repairs.} A repair maps estimates $\hat{\mathbf{y}}$ to values $\tilde{\mathbf{y}}$ that satisfy the constraint, $A\tilde{\mathbf{y}} = \mathbf{0}$. We compare two post-hoc repairs, which need no further model call, with an in-context alternative. \emph{Reconciliation by averaging}~\eqref{eq:avg} sets each tract value to the mean of its answer and its block-group sum, and scales the block groups to that total. The \emph{block-group sum}~\eqref{eq:sum}, the simplest reconciliation, takes the sum as the tract value and leaves the block groups unchanged:
\begin{align}
\tilde y_p &= \tfrac{1}{2}\bigl(\hat y_p + S_p\bigr), & \tilde y_c &= \hat y_c\,\tilde y_p / S_p, \label{eq:avg}\\
\tilde y_p &= S_p, & \tilde y_c &= \hat y_c. \label{eq:sum}
\end{align}
\emph{Joint prompting} lists a tract and all its block groups in one call and asks for every value: the nesting is shown, but no prompt asks for the values to add up, so satisfying the constraint is left to the model. Two further prompts supply partial structure: the \emph{count} prompt states how many block groups the tract has, and the \emph{anchor} prompt quotes the model's own tract answer in each block-group query.

\paragraph{Accuracy.} Study~1 scores answers against American Community Survey (ACS) estimates by median relative error (MdRE). Study~2 scores them against exact 2020 Census counts by median absolute log error (MdALE; a value of 0.18 corresponds to a typical answer off by a factor of 1.2).

\section{Experimental Design}\label{sec:design}

\paragraph{Models and settings.} A \emph{setting} is a model at one reasoning effort. The two main settings are frontier reasoning models at low reasoning effort: gpt-6.1-sol, at its default temperature, and grok-4.7, at temperature 0. Seven further settings test generality: gpt-6.1-sol at medium and high effort; gpt-6-luna without reasoning and at low effort; gpt-5.4 without reasoning; and, without reasoning, gpt-4.1 and the open-weights DeepSeek-V4-Pro. gpt-5.4, gpt-4.1 and DeepSeek-V4-Pro ran at temperature 0, and gpt-6-luna at its default temperature.

\subsection{Study 1: Nine Settings, 30 Tracts}\label{sec:s1design}

Cook County, Illinois, has 1,332 census tracts and 4,002 block groups~\cite{tiger2023}. We drew a stratified sample of 30 tracts, six from each income quintile, each with two to five block groups; with their 89 block groups this gives 119 areas. Reference values are ACS 2018--2022 five-year estimates~\cite{acs2022}. Their block-group population totals add up exactly to the tract totals, so the constraint holds for the reference values too; Study~2 scores accuracy against exact 2020 Census counts. Each prompt gives the type of area, its Census code, a description (for example, ``Block Group 1 of Census Tract 6609 in the Chicago Lawn community area of Chicago''), a point inside it and its land area, and asks for the ACS population and median household income, instructing the model to give its best estimate rather than refuse (Appendix~\ref{app:details}). The main settings answered all 119 questions three times and the others two or three times (Table~\ref{tab:s1settings}); results use answers averaged over runs unless stated. Nine block groups have no published income; they are excluded from every income comparison, for census values and models alike, which leaves 29 tracts. Confidence intervals resample tracts 1,000 times with their block groups. Three GPT settings (gpt-6.1-sol and gpt-6-luna at low effort, and gpt-5.4) also answered the joint, count and anchor prompts, three runs each, and grok-4.7 and DeepSeek-V4-Pro answered the joint prompt.

Two baselines calibrate the results. \emph{County density $\times$ area} multiplies Cook County's average density, about 2,156 people per km$^2$~\cite{census2020,gazetteer2023}, by each area's land area; it is crude but coherent by construction. \emph{Reference regressions} estimate one area at a time, like the models: log-log regressions of ACS population on the information in the prompt, fitted for each level on all other Cook County areas and made mean-unbiased by Duan smearing. Inputs are added in steps: typical size (each level's county median population), land area, location (log distance to the Loop in downtown Chicago, latitude and longitude) and community area or place.

\subsection{Study 2: Pre-specified Confirmation, 300 Tracts}\label{sec:s2design}

Study~2 tests the coherence findings at ten times the scale of Study~1, against exact counts. Its hypotheses, tests and exclusion rules were fixed in a written analysis plan before any model was queried (Appendix~\ref{app:s2}). The sample comprises 300 of Cook County's 1,332 tracts and their 964 block groups: all 49 tracts with six or seven block groups, 60 drawn at random from the 164 with a single block group, and 191 drawn in proportion from those with two to five. Answers are scored against the 2020 Census counts of population and housing units~\cite{census2020}, which add up exactly from block groups to tracts. Prompts describe each area as in Study~1 and request the two counts; a joint prompt lists a tract and all its block groups in one call. Three GPT settings were evaluated: gpt-6.1-sol and gpt-6-luna at low effort, and gpt-5.4 without reasoning. Each setting answered every area three times, individually and jointly, and the tract question for each held-out tract nine further times.

The plan specified four confirmatory predictions about the coherence error: that the coherence error of averaged answers exceeds the 97.5th percentile of noise alone; that it increases with $k$, the slope of $\ln(S_p / \hat y_p)$ on $\ln k$ being positive; that for $k = 1$ the block-group answer falls below the tract answer (Wilcoxon signed-rank test, Holm-corrected); and that joint answers add up. Table~\ref{tab:s2tests} reports the pre-specified tests. Tracts were randomly assigned, within each $k$, to a development half and a held-out half. A correction method developed on the development half was compared, on the held-out tracts, with simple comparators fixed in the plan, among them both reconciliations, the joint answers and the mean of 12 tract runs. Comparisons among these simple methods are exploratory. Intervals are paired bootstraps over tracts with 2,000 resamples.

\section{Results}\label{sec:results}

\subsection{Every Setting Violates the Constraint}\label{sec:r1}

\begin{figure*}[t]
\centering
\begin{tikzpicture}[
  box/.style={draw, rounded corners, align=center, font=\footnotesize, inner sep=3pt, minimum width=17mm},
  sum/.style={draw, rounded corners, fill=black!5, align=center, font=\footnotesize, inner sep=3pt},
  ttl/.style={font=\small\bfseries, align=center}]
  \begin{scope}
    \node[ttl] at (3.3,1.2) {gpt-6.1-sol: Tract 8046.06, Elk Grove Village};
    \node[box] (t1) at (3.3,0.35) {Tract\\6,349 / 5,420};
    \node[box] (g1) at (0.0,-1.05) {BG 1\\441 / 1,224};
    \node[box] (g2) at (2.2,-1.05) {BG 2\\2,456 / 1,347};
    \node[box] (g3) at (4.4,-1.05) {BG 3\\1,096 / 2,864};
    \node[box] (g4) at (6.6,-1.05) {BG 4\\2,356 / 2,240};
    \foreach \n in {g1,g2,g3,g4} \draw (t1.south) -- (\n.north);
    \node[sum] (s1) at (3.3,-2.5) {Block groups add up to 7,675:\\42\% above the model's tract answer};
    \foreach \n in {g1,g2,g3,g4} \draw[-{Stealth[length=1.5mm]}] (\n.south) -- (s1.north);
  \end{scope}
  \begin{scope}[xshift=9.6cm]
    \node[ttl] at (2.2,1.2) {grok-4.7: Tract 3510, Douglas};
    \node[box] (t2) at (2.2,0.35) {Tract\\3,526 / 2,145};
    \node[box] (k1) at (0.0,-1.05) {BG 1\\540 / 1,486};
    \node[box] (k2) at (2.2,-1.05) {BG 2\\917 / 1,094};
    \node[box] (k3) at (4.4,-1.05) {BG 3\\2,069 / 1,247};
    \foreach \n in {k1,k2,k3} \draw (t2.south) -- (\n.north);
    \node[sum] (s2) at (2.2,-2.5) {Block groups add up to 3,827:\\78\% above the model's tract answer};
    \foreach \n in {k1,k2,k3} \draw[-{Stealth[length=1.5mm]}] (\n.south) -- (s2.north);
  \end{scope}
\end{tikzpicture}
\caption{Study 1, run 1: the worst tract for each main model. The census block groups add up exactly; the model's do not. Each box shows the census value, then the model's estimate of population (BG = block group).}
\label{fig:tracts}
\end{figure*}
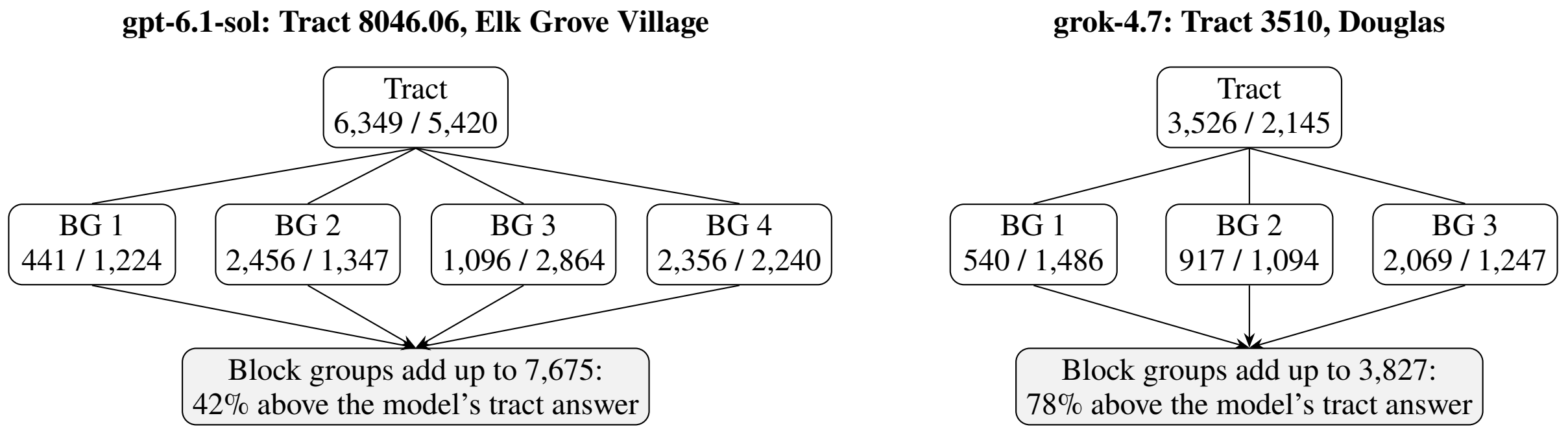
\begin{table*}[t]
\centering
{\footnotesize
\setlength{\tabcolsep}{4pt}
\begin{tabularx}{\textwidth}{@{}lcCCccC@{}}
\toprule
Setting & Runs / tracts & Coherence error, median [95\% CI] & Noise alone, average / single run & Tract error & Pull slope & Sum vs.\ tract, 2 / 5 block groups \\
\midrule
gpt-6.1-sol, low & 3 / 30 & 12.0\% [5.6, 21.6] & 4.9\% / 9.8\% & 15.0\% & 0.59 & $-$1\% / +33\% \\
gpt-6.1-sol, medium & 2 / 30 & 12.6\% [8.2, 16.8] & 4.7\% / 8.7\% & 19.1\% & 0.64 & $-$8\% / +57\% \\
gpt-6.1-sol, high & 2 / 30 & 13.4\% [8.2, 24.2] & 4.4\% / 7.4\% & 14.7\% & 0.62 & $-$2\% / +20\% \\
gpt-6-luna, none & 3 / 28 & 16.2\% [13.8, 22.4] & 5.0\% / 9.0\% & 19.5\% & 0.16 & $-$18\% / +35\% \\
gpt-6-luna, low & 3 / 30 & 17.2\% [12.5, 19.3] & 6.7\% / 12.7\% & 22.3\% & 0.46 & $-$17\% / +27\% \\
gpt-5.4, none & 3 / 30 & 15.3\% [9.5, 28.2] & 6.7\% / 12.5\% & 16.7\% & 0.28 & $-$27\% / +42\% \\
gpt-4.1 & 2 / 30 & 21.4\% [12.5, 35.4] & 7.7\% / 15.2\% & 25.4\% & 0.25 & $-$24\% / +62\% \\
grok-4.7, low & 3 / 30 & 12.6\% [7.8, 17.5] & 5.2\% / 9.8\% & 16.7\% & 0.35 & $-$13\% / +34\% \\
DeepSeek-V4-Pro & 3 / 30 & 18.4\% [15.0, 24.8] & 3.4\% / 6.5\% & 23.1\% & 0.21 & $-$13\% / +65\% \\
\midrule
\emph{Regression:} typical size & -- / 30 & 27.3\% [27.3, 36.4] & -- & 33.5\% & 0.00 & $-$36\% / +59\% \\
\quad + land area & -- / 30 & 16.2\% [10.8, 21.3] & -- & 15.0\% & 0.19 & $-$21\% / +54\% \\
\quad + location & -- / 30 & 12.3\% [9.1, 21.0] & -- & 23.0\% & 0.28 & $-$15\% / +44\% \\
\quad + community area / place & -- / 30 & 16.8\% [10.1, 23.1] & -- & 17.4\% & 0.32 & $-$14\% / +52\% \\
\bottomrule
\end{tabularx}\par}
\caption{Study 1: every setting's coherence error exceeds what its run-to-run noise alone would produce. Nine settings and four reference regressions, each estimating one area at a time; model answers are averaged over runs. Noise alone: average of runs / single run, resampling model (Appendix~\ref{app:noise}). Tract error: MdRE against ACS. Pull slope: log block-group answer on log 2020 count (1 = no pull). Last column: signed coherence error in tracts with two and five block groups (12 and 2 tracts).}
\label{tab:s1settings}
\end{table*}

Neither main model's block-group answers add up to its own tract answers (Table~\ref{tab:s1settings}). Averaged over three runs, the median coherence error is 12.0\% for gpt-6.1-sol and 12.6\% for grok-4.7, and it exceeds 10\% in more than half the tracts. Single answers vary: answers for the same area differed between two runs by a median of 12.0--13.6\%, even at temperature 0. Yet a perfectly coherent model with this noise would miss, on the average of three runs, by only 4.8--5.9\% (gpt-6.1-sol) and 4.8--6.4\% (grok-4.7), depending on the noise model, and in none of 1,000 simulations under any of three noise models did it reach the observed values (Appendix~\ref{app:noise}). A single run, which is what a user sees, is no better: 12\% and 17\% in run 1. In each model's worst tract, the block-group answers add up to 42\% and 78\% more than the model's own tract answer, while the census block groups add up exactly (Figure~\ref{fig:tracts}).

The seven further settings behave alike: in every setting the averaged answers miss by more than noise alone allows. More reasoning makes gpt-6.1-sol's answers more repeatable (run-to-run disagreement on tracts of 12.0\%, 10.0\% and 8.5\% at low, medium and high effort) but not more coherent (12.0\%, 12.6\% and 13.4\%). Temperature 0 does not make answers repeat: gpt-5.4, gpt-4.1, grok-4.7 and DeepSeek-V4-Pro gave identical population answers in all runs for only 13\%, 24\%, 0\% and 22\% of areas.

\paragraph{Coherence is not accuracy.} Both main models are far more accurate than density $\times$ area (median tract error 15.0\% and 16.7\% against 62\%) and rank tracts better than land area alone (Spearman $\rho$ 0.87 and 0.84 against 0.79; Appendix~\ref{app:details}). The crude baseline adds up by construction yet is badly wrong; Section~\ref{sec:repair} returns to how coherence and accuracy relate.

\paragraph{Conclusions shift with the level.} Both main models rank the 29 tracts' incomes almost identically at the two levels (Spearman $\rho$ 0.97 and 0.96), yet their poorest ten change. Averaged over three runs, each model's lists from its tract answers and from its block-group answers overlap by 0.82, so one tract of the ten differs; two lists of ten drawn at random would overlap by about 0.21. In the census the list is the same at both levels, so the change is a spurious scale effect in a conclusion, not only in a total.

\subsection{Per-area Shrinkage Explains the Violation}\label{sec:why}

\paragraph{The answers are estimates, not recalled figures.} Recalled figures would add up; estimates need not. No tract answer exactly matched any of four published figures for its area (three ACS estimates and the 2020 count), and only one block-group answer did. Nor do the answers lean toward the 2020 count: they are closer to it than to the ACS figure asked for in 60\% (gpt-6.1-sol) and 61\% (grok-4.7) of areas, within three points of the 60\% and 58\% that an equally accurate estimator without recall would give (Appendix~\ref{app:recall}).

\paragraph{Answers are pulled toward a typical size.} Across the 89 block groups, a tenfold difference in population (the 2020 count) becomes about a fourfold difference in gpt-6.1-sol's averaged answers and a 2.2-fold difference in grok-4.7's. The \emph{pull slopes}, of log answer on log count, are 0.59 [0.37, 0.79] and 0.35 [0.20, 0.49], where 1 would mean no pull. This is shrinkage, the sound response of an estimator that is uncertain about an area, and it is not excessive: the \emph{calibration slope}, of log count on log answer, which exceeds 1 when answers are pulled too far, is 0.63 [0.46, 0.81] for gpt-6.1-sol and 0.80 [0.56, 1.04] for grok-4.7, and lies below 1 in eight of the nine settings.

\paragraph{Shrinking each area separately breaks the sum.} If every block group receives a similar answer, the block-group sum grows with $k$ while the tract answer does not. Replacing each block-group answer with the fitted pull line reproduces the overshoot in tracts with four or five block groups: +20\% and +33\% observed against +20\% and +35\% predicted for gpt-6.1-sol (Appendix~\ref{app:details}).

\paragraph{Regressions that estimate one area at a time do the same.} Given only each level's typical size, the reference regressions have a coherence error of 27.3\%; given land area, location or community area as well, 12.3--16.8\%, with the signature most settings show: a shortfall in tracts with two block groups and an excess in those with five ($-$15\% and +44\% with location; Table~\ref{tab:s1settings}). The best model settings (12.0--13.4\%) are about as coherent as the regressions, with similar tract accuracy (median error 14.7--19.1\% against 15.0--23.0\%). Over all 1,114 Cook County tracts with two to five populated block groups and a published income, the regressions' coherence error is 17.8--28.2\% at our mix of block-group counts. Any estimator that shrinks each area toward a typical size, without knowing how many block groups a tract has, leaves this signature.

\subsection{The Violation Replicates at Ten Times the Scale}\label{sec:r2}

\begin{table*}[t]
\centering
{\footnotesize
\begin{tabularx}{\textwidth}{@{}p{5.6cm}CCC@{}}
\toprule
 & gpt-6.1-sol, low & gpt-6-luna, low & gpt-5.4, none \\
\midrule
Coherence error, median & 16.0 / 16.2\% & 17.7 / 17.6\% & 27.2 / 28.6\% \\
Noise alone, 97.5th percentile & 6.0 / 6.0\% & 8.3 / 8.6\% & 11.2 / 11.2\% \\
Coherence error, 2--5 block groups$^\dagger$ & 12.1 / 11.7\% & 12.5 / 13.8\% & 20.3 / 21.3\% \\
\midrule
\multicolumn{4}{@{}l}{\emph{Signed coherence error by number of block groups $k$ (tracts)}} \\
$k = 1$, same area (57--60) & $-$17 / $-$17\% & $-$27 / $-$27\% & $-$29 / $-$29\% \\
$k = 2$ (61) & $-$2 / $-$4\% & $-$12 / $-$14\% & $-$7 / $-$8\% \\
$k = 3$ (59) & +9 / +7\% & +1 / +3\% & +10 / +11\% \\
$k = 4$ (52) & +21 / +21\% & +11 / +12\% & +8 / +10\% \\
$k = 5$ (19) & +29 / +28\% & +16 / +20\% & +30 / +36\% \\
$k = 6$--7 (49) & +37 / +33\% & +24 / +27\% & +44 / +49\% \\
\midrule
Joint call, coherence error & 0.0 / 0.0\% & 0.0 / 0.0\% & 0.0 / 0.0\% \\
Pull slope & 0.39 / 0.33 & 0.40 / 0.32 & 0.34 / 0.27 \\
Calibration slope & 0.56 / 0.46 & 0.39 / 0.37 & 0.31 / 0.30 \\
\bottomrule
\end{tabularx}\par}
\caption{Study 2: the coherence error replicates at ten times the scale and increases with the number of block groups. Population / housing units, answers averaged over three runs, 300 tracts. Noise alone: resampling model (Appendix~\ref{app:noise}), 1,000 simulations. Signed coherence error: median of (sum $-$ tract answer) / tract answer; at $k = 1$ the block group and the tract are the same area. Slopes as in Section~\ref{sec:why}, against 2020 counts. $^\dagger$Study~1's range of $k$; not a pre-specified test.}
\label{tab:s2}
\end{table*}

\begin{table*}[t]
\centering
{\small
\setlength{\tabcolsep}{4pt}
\begin{tabular}{@{}lcccccc@{}}
\toprule
& \multicolumn{3}{c}{Tract} & \multicolumn{3}{c}{Block group} \\
\cmidrule(lr){2-4}\cmidrule(lr){5-7}
Setting & As asked & Reconciled & Joint & As asked & Reconciled & Joint \\
\midrule
gpt-6.1-sol, low      & 15.0\% & 11.0\% ($-$4.0) & 14.2\% & 28.4\% & 28.1\% ($-$0.3) & 26.9\% \\
gpt-6.1-sol, medium   & 19.1\% & 13.1\% ($-$6.0)$^*$ & --   & 31.1\% & 28.8\% ($-$2.4) & -- \\
gpt-6.1-sol, high     & 14.7\% & 10.4\% ($-$4.3)$^*$ & --   & 27.5\% & 27.6\% (+0.2) & -- \\
gpt-6-luna, none      & 20.2\% & 15.8\% ($-$4.4) & --     & 29.7\% & 26.3\% ($-$3.4) & -- \\
gpt-6-luna, low       & 22.3\% & 17.9\% ($-$4.4) & 20.4\% & 30.0\% & 27.5\% ($-$2.6) & 35.1\% \\
gpt-5.4, none         & 16.7\% & 10.5\% ($-$6.2)$^*$ & 22.0\% & 29.7\% & 30.0\% (+0.3) & 38.2\% \\
gpt-4.1               & 25.4\% & 21.5\% ($-$4.0) & --     & 30.7\% & 33.5\% (+2.8) & -- \\
grok-4.7, low         & 16.7\% & 13.8\% ($-$2.9) & 18.8\% & 32.2\% & 31.0\% ($-$1.2) & 36.1\% \\
DeepSeek-V4-Pro       & 23.1\% & 18.5\% ($-$4.6) & 21.3\% & 31.3\% & 30.1\% ($-$1.3) & 38.5\% \\
\midrule
Regression + place, 30 tracts & 17.4\% & 10.7\% ($-$6.7) & -- & 27.4\% & 28.1\% (+0.7) & -- \\
Regression + place, all Cook  & 20.3\% & 16.7\% ($-$3.7)$^*$ & -- & 27.0\% & 26.7\% ($-$0.3) & -- \\
\bottomrule
\end{tabular}\par}
\caption{Study 1: reconciling after generation lowers tract error in every setting. Median relative error against ACS of averaged answers; reconciliation and joint prompting as defined in Section~\ref{sec:method}. Changes against \emph{as asked} in percentage points; $^*$the 95\% paired bootstrap interval over tracts excludes 0. gpt-6-luna without reasoning: 28 tracts; others 30.}
\label{tab:s1reconcile}
\end{table*}

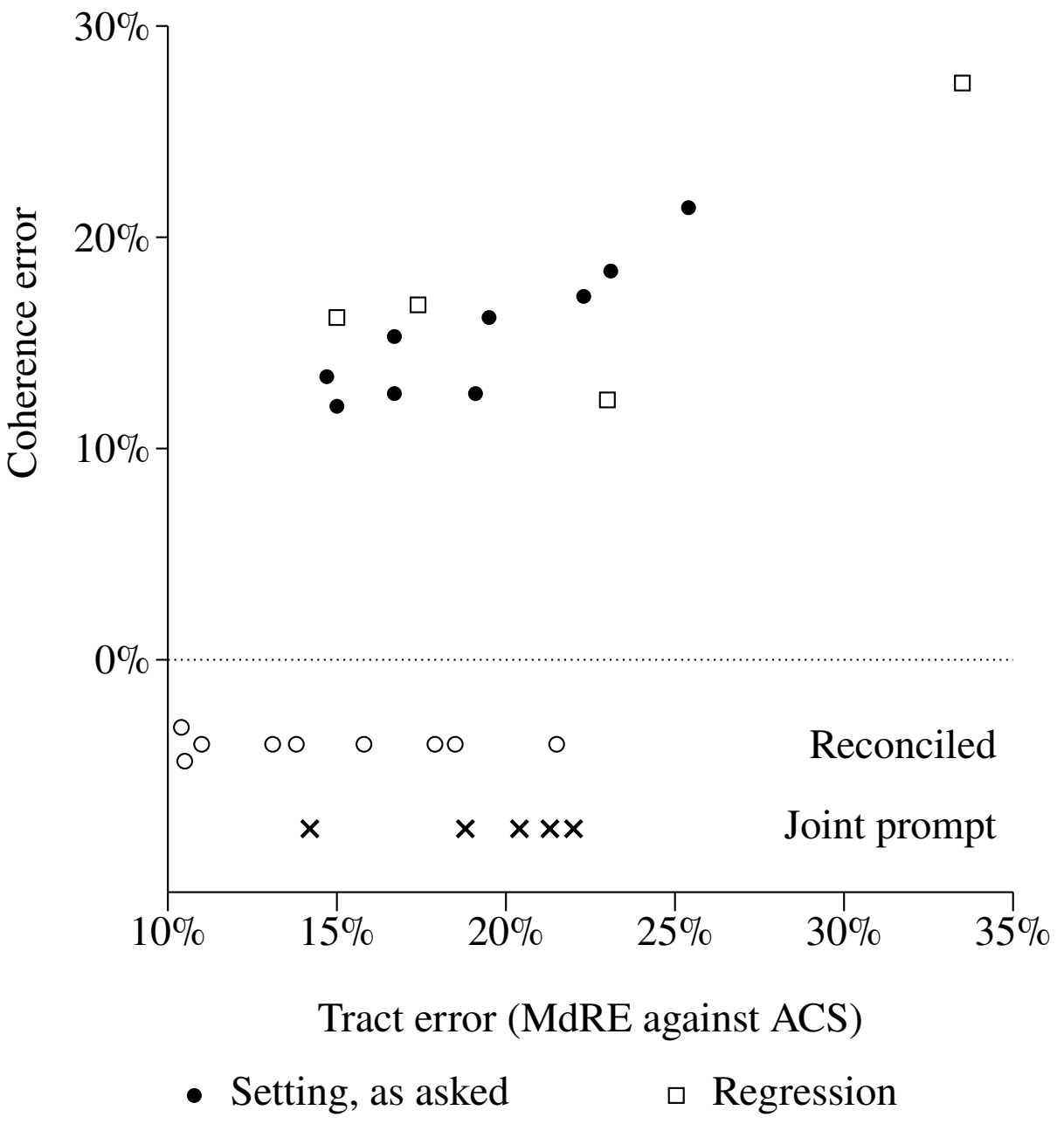
\begin{figure}[t]
\centering
\begin{tikzpicture}[x=0.256cm, y=0.16cm, font=\footnotesize]
  \draw (0,-11) -- (25,-11);
  \draw (0,-11) -- (0,30);
  \draw[densely dotted] (0,0) -- (25,0);
  \foreach \xv in {10,15,20,25,30,35} \draw ({\xv-10},-11) -- ({\xv-10},-11.9) node[below, inner sep=1pt] {\xv\%};
  \foreach \yv in {0,10,20,30} \draw (0,\yv) -- (-0.35,\yv) node[left, inner sep=1pt] {\yv\%};
  \node[anchor=north] at (12.5,-15.5) {Tract error (MdRE against ACS)};
  \node[rotate=90, anchor=south] at (-3.4,15) {Coherence error};
  \foreach \xv/\yv in {15.0/12.0, 19.1/12.6, 14.7/13.4, 19.5/16.2, 22.3/17.2, 16.7/15.3, 25.4/21.4, 16.7/12.6, 23.1/18.4}
    \fill ({\xv-10},\yv) circle (1.6pt);
  \foreach \xv/\yv in {33.5/27.3, 15.0/16.2, 23.0/12.3, 17.4/16.8}
    \draw ({\xv-10},\yv) ++(-1.6pt,-1.6pt) rectangle ++(3.2pt,3.2pt);
  \foreach \xv/\yv in {11.0/-4, 13.1/-4, 10.4/-3.2, 15.8/-4, 17.9/-4, 10.5/-4.8, 21.5/-4, 13.8/-4, 18.5/-4}
    \filldraw[fill=white] ({\xv-10},\yv) circle (1.6pt);
  \node[anchor=east] at (25,-4) {Reconciled};
  \foreach \xv in {14.2, 20.4, 22.0, 18.8, 21.3}
    \draw[thick] ({\xv-10},-8) ++(-1.8pt,-1.8pt) -- ++(3.6pt,3.6pt) ++(0,-3.6pt) -- ++(-3.6pt,3.6pt);
  \node[anchor=east] at (25,-8) {Joint prompt};
  \fill (0.2cm,-3.3cm) circle (1.6pt);
  \node[anchor=west] at (0.35cm,-3.3cm) {Setting, as asked};
  \draw (3.85cm,-3.3cm) ++(-1.6pt,-1.6pt) rectangle ++(3.2pt,3.2pt);
  \node[anchor=west] at (4.0cm,-3.3cm) {Regression};
\end{tikzpicture}
\caption{Study 1: coherence error rises with tract error across the nine settings (Spearman $\rho$ 0.83). Reconciliation brings every setting's coherence error to zero and lowers its tract error; joint prompting also satisfies the constraint, with higher tract error than reconciliation in all five settings. Coherent answers are shown in the strip below the dotted zero line. Points from Tables~\ref{tab:s1settings} and~\ref{tab:s1reconcile}.}
\label{fig:coherence}
\end{figure}

All four pre-specified predictions about the coherence error were supported in all three settings and for both variables (Table~\ref{tab:s2}; Appendix~\ref{app:s2} reports every test). The median coherence error of averaged answers, 16.0--28.6\%, is more than twice the 97.5th percentile of noise alone, 6.0--11.2\%. Restricted to the 191 tracts with two to five block groups, the range sampled in Study~1, gpt-6.1-sol's median coherence error is 12.1\%, against 12.0\% in Study~1. Joint answers were coherent (median coherence error 0.0\%) without splitting the tract evenly: near-even splits accounted for only 4.5--8.4\% of joint answers, and within-tract rankings did not deteriorate (gpt-5.4's improved).

\paragraph{The error increases with the number of block groups.} As per-area shrinkage predicts, the block-group sum falls short of the tract answer in tracts with one or two block groups and exceeds it beyond that, increasingly with $k$, by +24\% to +44\% for population in the 49 tracts with six or seven (slope on $\ln k$ 0.29--0.37; all intervals exclude 0).

\paragraph{Same place, different answer.} In the 60 sampled tracts that contain a single block group, the tract and the block group are the same area, with the same centre point, land area and published counts; the prompts differ only in the area type, the code and the description. Yet the block-group answer is lower by a median of 17\% (gpt-6.1-sol), 27\% (gpt-6-luna) and 29\% (gpt-5.4), for population and housing units alike (Holm-corrected $p < 10^{-7}$ in all six tests). In each model, the answer under one of the two labels is biased toward that level's typical size. gpt-6.1-sol's and gpt-6-luna's tract answers are close to the count, but their block-group population answers, 1,228 and 1,220, match the county's median block group (1,218) rather than these areas' median count (1,526); gpt-5.4's block-group answers are close to the count, whereas its tract answers are biased upward (median log error +0.45). A model that recalled the published count would give both questions the same answer. The models can reproduce published figures (all three returned Cook County's exact 2020 population in every run), yet at tract and block-group level almost no answer matches the count: at most 3 of about 890 tract answers and 14 of about 2,880 block-group answers per setting and variable.

\subsection{The Constraint as Verifier and Repair Operator}\label{sec:repair}

\begin{table*}[t]
\centering
{\footnotesize
\begin{tabularx}{\textwidth}{@{}p{5.6cm}CCC@{}}
\toprule
 & gpt-6.1-sol, low & gpt-6-luna, low & gpt-5.4, none \\
\midrule
\multicolumn{4}{@{}l}{\emph{Tracts}} \\
As asked & 0.180 / 0.233 & 0.276 / 0.255 & 0.292 / 0.290 \\
Mean of 12 runs & 0.178 / 0.224 & 0.263 / 0.247 & 0.264 / 0.245 \\
Joint call & 0.175 / 0.168 & 0.256 / 0.252 & 0.280 / 0.231 \\
Reconciled (averaged) & 0.144 / 0.173 & 0.200 / 0.220 & 0.189 / 0.212 \\
Block-group sum & 0.129 / 0.134 & 0.184 / 0.198 & 0.204 / 0.219 \\
Count only ($k$ $\times$ county answer / 4,002) & 0.221 / 0.273 & 0.221 / 0.296 & 0.224 / 0.271 \\
\multicolumn{4}{@{}l}{\emph{Block groups}} \\
As asked (kept by the block-group sum) & 0.179 / 0.192 & 0.278 / 0.269 & 0.268 / 0.288 \\
Joint call & 0.235 / 0.226 & 0.355 / 0.352 & 0.302 / 0.308 \\
Reconciled (scaled to the averaged total) & 0.208 / 0.230 & 0.286 / 0.287 & 0.270 / 0.283 \\
\bottomrule
\end{tabularx}\par}
\caption{Study 2, held-out tracts (exploratory comparison): the block-group sum outperforms the tract answer asked once, asked 12 times or asked jointly, and a rule that uses only the number of block groups, matches averaging, and leaves the block-group answers unchanged. Median absolute log error against 2020 counts, population / housing units, 144--145 tracts and 476--477 block groups per setting, every method scored on the same areas. Mean of 12 runs: the tract asked 12 times. Count only: the tract's number of block groups $k$ times the model's own county answer per block group (Cook County has 4,002).}
\label{tab:s2fix}
\end{table*}

\paragraph{Coherence error tracks tract error across settings.} Across the nine settings of Study~1, a setting's coherence error tracks its tract error (Spearman $\rho$ 0.83, exact $p = 0.008$; Figure~\ref{fig:coherence}), as consistency tracks accuracy for LLM forecasters~\cite{paleka2024}; with the four regressions added, $\rho$ is 0.63 ($n = 13$). Because the check needs no answer key, it offers a way to compare settings where no published data exist. For grok-4.7 and gpt-5.4, a 10\% coherence-error threshold also works within a setting: it catches 9 of 10 tracts more than 20\% off for grok-4.7 (against 6.0 expected by chance; $p = 0.021$) and 10 of 11 for gpt-5.4 ($p = 0.037$).

\paragraph{Reconciliation repairs the answers without another call.} Averaging each tract answer with its block-group sum, and scaling the block groups to that total, makes the levels add up by construction. It lowered tract error in all nine settings, by 2.9 to 6.2 points, and left block-group error within 3.4 points of the answers as asked (Table~\ref{tab:s1reconcile}). For estimates shrunk toward a typical size, the sum of the parts is a useful second estimate of the whole: the same reconciliation lowers the regressions' tract error across all Cook County tracts from 20.3\% to 16.7\%.

\paragraph{Joint prompting satisfies the constraint but costs block-group accuracy.} Asked jointly, all five settings returned block groups that add up to the tract within 1\% in every tract and run, except in 1\% of DeepSeek-V4-Pro's tracts and runs. The models achieve this by choosing a tract total and dividing it, in proportion to land area in 11--80\% of answers and evenly in only 7--14\%. On Study~1's tracts, gpt-6.1-sol kept its accuracy, while the other four settings, from three vendors, traded block-group accuracy for coherence; in no setting were the joint block groups more accurate than an even split of the model's own joint tract answer. Reconciliation was the better repair: its tracts were more accurate than the joint ones in all five settings, and its block groups in four. Partial structure is not enough. Stating the number of block groups lowered the coherence error for two of three settings (gpt-6.1-sol from 12.0\% to 9.7\%, gpt-6-luna from 17.2\% to 12.0\%), and quoting the model's own tract answer helped none: gpt-5.4 divided the quoted total by about three whatever the tract (Appendix~\ref{app:details}).

\paragraph{At scale, the block-group sum is the simplest effective repair.} In an exploratory comparison on Study~2's held-out tracts (Table~\ref{tab:s2fix}), reconciliation by averaging lowered tract error in all six setting $\times$ variable cells (gpt-6.1-sol population: from 0.180 to 0.144; five of six intervals exclude 0), but scaling the block groups to the averaged total increased their error in five, significantly only for gpt-6.1-sol (0.179 to 0.208). Taking the block-group sum as the tract answer avoids that cost: it lowered tract error in all six cells by about as much as averaging (0.129) and leaves the block-group answers unchanged. With about ten calls per tract, it also outperformed the mean of 12 tract runs in every cell (0.178; intervals exclude 0 in four of six). The block-group answers add information beyond the number of block groups: a count-only rule, $k$ times the model's own county answer per block group, had higher tract error than the block-group sum in all six cells (0.221; intervals exclude 0 in four of six), although for gpt-6-luna's population and both of gpt-5.4's variables it was more accurate than the tract answers as asked. This parallels the macro fallacy that Wolf et al.\ report for persona subpopulations~\cite{wolf2026partition}, here for physical areas with exact counts and against a count-only control. Joint answers had lower tract error than the answers as asked but higher block-group error, in all six cells (0.235), and the block-group sum outperformed joint prompting at both levels in every cell. The plan's projection-based reconciliations, standard in forecasting~\cite{hyndman2011,wickramasuriya2019}, did not do better overall: ordinary least-squares projection had higher error than the block-group sum at both levels in all six cells, and the weighted projection matched averaging for tracts. Even the correction designed for this task and fixed in the plan, which anchors each level on the model's own county answer, did not reliably outperform these simple methods.

\section{Discussion}\label{sec:discussion}

\paragraph{The census separates the model's scale effect from the data's.} For counts, the real gap between nested levels is zero by definition, as is the real difference between a tract and its only block group; for the tracts of Study~1, the real effect on the poorest-ten list is also zero. Every scale effect we observe is therefore produced by the models. In maps and rankings such effects can pass for real findings: run with the same model one area at a time, the scale sensitivity check that a recent governance review recommends~\cite{governance} could report a scale effect that is not in the data.

\paragraph{Enforce constraints outside the model.} The additive constraint is cheap to check, needs no answer key and, across settings, tracks tract error. Enforcing it after generation lowered tract error in every setting, whereas joint prompting, which shows the nesting but leaves the constraint to the model, satisfied it by apportioning a chosen total and usually degraded the block groups. Reconciliation gave more accurate tracts than joint prompting in all five settings of Study~1, and the block-group sum did so in all six cells of Study~2 (exploratory), consistent with results for probabilistic claims composed by LLM agents, which projection repairs and prompting-side fixes do not~\cite{kotawala2026locally}. This suggests a general recipe for structured estimation with foundation models: query the parts, check the constraints, reconcile the levels~\cite{wickramasuriya2019,wang2008sae}, and check accuracy wherever an answer key exists.

\paragraph{Generality.} The test requires only a hierarchy with additive counts, so it applies directly to other regions, to other variables such as jobs, and to alternative zonings such as hexagon grids and random regions~\cite{nieves2024}.

\section{Conclusion}\label{sec:conclusion}

Census hierarchies give LLM area estimates an exact, label-free consistency test. All nine settings failed it: block-group estimates did not add up to the models' own tract estimates beyond what run-to-run noise allows, and a pre-specified confirmatory study at ten times the scale replicated the failure and showed the same area estimated 17--29\% lower when presented as a block group. Per-area shrinkage toward a typical size reproduces the coherence error. The constraint that exposes these effects also repairs them: reconciliation after generation made the answers coherent and lowered tract error in every setting, while joint prompting, which leaves the constraint to the model, usually cost block-group accuracy.

\bibliography{references}

\appendix
\section{Study 1 Details}\label{app:details}

\paragraph{Prompt.} The full prompt for one tract follows; block-group prompts differ only in the area type, the code and the description. Joint prompts list the tract and each of its block groups in the same format and ask for every value; count prompts add the number of block groups in the tract, or state that the block group is one of $N$ in its tract; anchor prompts quote the model's own tract answer from the same baseline run.
\begin{lstlisting}[frame=tb]
Below is a US Census Bureau geographic area in Cook County, Illinois.

Area type: census tract
GEOID: 17031660900
Description: Census Tract 6609 in the Chicago Lawn community area of Chicago, Cook County, Illinois
Centre (Census internal point, WGS84 latitude, longitude): 41.767819, -87.699063
Land area: 1.811 km² (0.699 sq mi)

Estimate the following American Community Survey (ACS) 2018-2022 5-year values for this exact area:
1. Total population (table B01003).
2. Median household income in the past 12 months, in 2022 inflation-adjusted US dollars (table B19013).

Give your best numeric estimates even if you are uncertain; do not refuse. End your reply with exactly these two lines:
POPULATION: <integer>
MEDIAN_INCOME: <integer>
\end{lstlisting}

\begin{table*}[t]
\centering
{\small
\begin{tabularx}{\textwidth}{@{}p{3.6cm}CCCC@{}}
\toprule
Source & Coherence error, median [95\% CI] & Tracts with coherence error above 10\% & Poorest-ten overlap, tract vs.\ block group & Tract-level list vs.\ census poorest ten \\
\midrule
Census & 0\% & 0\% & 1.00 & 1.00 \\
County density $\times$ area & 0\% & 0\% & n/a (no income) & n/a \\
gpt-6.1-sol, mean of 3 runs & 12.0\% [5.6, 21.6] & 53\% & 0.82 [0.54, 1.00] & 0.54 \\
gpt-6.1-sol, run 1 & 12\% [7, 18] & 57\% & 0.67 [0.54, 1.00] & 0.54 \\
grok-4.7, mean of 3 runs & 12.6\% [7.8, 17.5] & 60\% & 0.82 [0.67, 1.00] & 0.54 \\
grok-4.7, run 1 & 17\% [7, 24] & 63\% & 0.82 [0.67, 1.00] & 0.54 \\
\bottomrule
\end{tabularx}\par}
\caption{Study 1, main models: the answers do not fit together across levels; the census values do. Coherence error uses all 30 tracts; the poorest-ten overlaps use the 29 tracts with published block-group income, for the census and the models alike. Model rows are averaged over three runs or from run 1. Brackets: 95\% bootstrap intervals.}
\label{tab:s1coherence}
\end{table*}

\begin{table}[t]
\centering
{\footnotesize
\setlength{\tabcolsep}{3pt}
\begin{tabularx}{\columnwidth}{@{}llCCC@{}}
\toprule
Level & Variable & gpt-6.1-sol & grok-4.7 & Density $\times$ area \\
\midrule
Tract & Population & 0.87 / 15.0\% & 0.84 / 16.7\% & 0.79 / 62\% \\
Tract & Income & 0.74 / 17.7\% & 0.76 / 24.4\% & -- \\
BG & Population & 0.54 / 28.4\% & 0.44 / 32.2\% & 0.46 / 60\% \\
BG & Income & 0.60 / 26.1\% & 0.56 / 29.0\% & -- \\
\bottomrule
\end{tabularx}\par}
\caption{Study 1, accuracy against the census: both models' errors are far below those of density $\times$ area. Spearman rank correlation ($\rho$) / median relative error (MdRE). Model columns are answers averaged over three runs.}
\label{tab:s1accuracy}
\end{table}

\begin{table}[t]
\centering
{\footnotesize
\setlength{\tabcolsep}{3pt}
\begin{tabularx}{\columnwidth}{@{}ccCCCC@{}}
\toprule
 & & \multicolumn{2}{c}{gpt-6.1-sol} & \multicolumn{2}{c}{grok-4.7} \\
\cmidrule(lr){3-4}\cmidrule(l){5-6}
$k$ & Tracts & Observed & Pull only & Observed & Pull only \\
\midrule
2 & 12 & $-$1\% & +1\% & $-$13\% & $-$1\% \\
3 & 9 & +1\% & $-$8\% & $-$6\% & $-$10\% \\
4 & 7 & +20\% & +20\% & +10\% & +10\% \\
5 & 2 & +33\% & +35\% & +34\% & +43\% \\
\bottomrule
\end{tabularx}\par}
\caption{Study 1: the block-group sum overshoots the model's own tract answer in tracts with four or five block groups, as the pull alone predicts. Signed coherence error by number of block groups $k$, for answers averaged over three runs: median of (sum $-$ tract answer) / tract answer. ``Pull only'' replaces each block-group answer with the fitted pull line and keeps the tract answer.}
\label{tab:s1bycount}
\end{table}

\begin{table*}[t]
\centering
{\footnotesize
\begin{tabularx}{\textwidth}{@{}p{4.4cm}CCCCC@{}}
\toprule
Condition & gpt-6.1-sol, low & gpt-6-luna, low & gpt-5.4, none & grok-4.7, low & DeepSeek-V4-Pro \\
\midrule
Baseline, one area at a time & 12.0\% & 17.2\% & 15.3\% & 12.6\% & 18.4\% \\
Count: number of block groups given & 9.7\% [5.4, 15.1] & 12.0\% [8.4, 17.1] & 19.8\% [11.1, 29.0] & -- & -- \\
Joint: tract and block groups in one call & 0.0\% & 0.0\% & 0.0\% & 0.0\% & 0.0\% \\
Anchor: own tract answer given & 14.6\% [9.4, 17.4] & 21.6\% [11.4, 28.9] & 30.7\% [19.6, 32.6] & -- & -- \\
\bottomrule
\end{tabularx}\par}
\caption{Study 1, asking with the nesting given. Only the joint call makes the answers add up in every setting. Median coherence error of answers averaged over three runs [95\% CI]; Table~\ref{tab:s1reconcile} gives the accuracy of the joint answers. grok-4.7 and DeepSeek-V4-Pro were asked jointly only.}
\label{tab:s1nested}
\end{table*}

\paragraph{Main models and income.} Table~\ref{tab:s1coherence} gives the coherence error and the poorest-ten overlaps of the two main models, averaged over three runs and for run 1; Table~\ref{tab:s1accuracy} gives their accuracy against the census and against density $\times$ area; Table~\ref{tab:s1bycount} gives the signed coherence error by number of block groups, with the prediction of the pull line alone.

\paragraph{Joint answers.} Joint answers add up because the models choose a tract total and divide it (gpt-5.4: it ``apportioned population primarily by land area''), in proportion to land area in 11--80\% of answers (DeepSeek-V4-Pro 80\%, grok-4.7 56\%) and evenly in only 7--14\%. In no setting are the joint block groups more accurate than an even split of the model's own joint tract answer (gpt-5.4: 38.2\% against 30.8\%; gpt-6.1-sol: 26.9\% against 25.7\%).

\paragraph{Count and anchor prompts.} Being told the count lowered the coherence error for gpt-6.1-sol and gpt-6-luna (12.0\% to 9.7\% and 17.2\% to 12.0\%) and flattened their pattern across block-group counts; gpt-5.4's rose from 15.3\% to 19.8\% (Table~\ref{tab:s1nested}). Being given its own tract answer helped none of them: separate block-group calls still did not add up to it. The anchor prompt quotes the tract answer but not how many block groups the tract has. gpt-5.4 divided the quoted total by about three whatever the tract, and gpt-6-luna nearly so, so their sums fell about a third short for two block groups and overshot for five; gpt-6.1-sol's divisor grew with the count, and it missed by less.

\begin{table*}[tp]
\centering
{\footnotesize
\begin{tabular}{@{}lcccccc@{}}
\toprule
Area & Census population & gpt-6.1-sol & grok-4.7 & Census income & gpt-6.1-sol & grok-4.7 \\
\midrule
Tract 6609 & 5,631 & 3,218 & 5,821 & \$41,032 & \$45,760 & \$41,250 \\
Block group 1 & 884 & 1,050 & 1,350 & not published & \$41,250 & \$42,800 \\
Block group 2 & 2,065 & 1,057 & 1,254 & \$56,840 & \$46,250 & \$39,250 \\
Block group 3 & 2,682 & 1,247 & 1,842 & \$58,625 & \$47,600 & \$43,650 \\
Sum of block groups & 5,631 & 3,354 & 4,446 & & & \\
\bottomrule
\end{tabular}\par}
\caption{Study 1, Census Tract 6609 (Chicago Lawn). Answers that nearly add up can still be far off.}
\label{tab:case1}
\end{table*}

\paragraph{Adding up is not being right.} Census Tract 6609 shows that coherence and accuracy are different properties (Table~\ref{tab:case1}). gpt-6.1-sol's answers nearly fit together, with a coherence error of 4.2\%, but both totals are about 40\% below the census value. grok-4.7's tract answer is within 4\% of the census value, but its block groups add up to 24\% less than its own tract answer.

\section{Coherence Error from Noise Alone}\label{app:noise}

\begin{table*}[tp]
\centering
{\footnotesize
\begin{tabularx}{\textwidth}{@{}LCCCC@{}}
\toprule
Noise model & gpt-6.1-sol, single run & gpt-6.1-sol, average of 3 & grok-4.7, single run & grok-4.7, average of 3 \\
\midrule
Resampling & 9.8\% [6.5, 13.3] & 4.9\% [3.0, 7.0] & 9.8\% [6.6, 13.7] & 5.2\% [3.1, 7.4] \\
Own variance & 8.4\% [5.0, 12.0] & 4.8\% [2.9, 6.9] & 8.3\% [5.3, 12.2] & 4.8\% [3.0, 7.1] \\
One variance & 10.2\% [6.4, 14.9] & 5.9\% [3.7, 8.6] & 11.1\% [7.2, 16.0] & 6.4\% [4.0, 9.1] \\
Observed & 14.6\% (mean of runs) & 12.0\% & 16.6\% (mean of runs) & 12.6\% \\
Share of simulations at or above observed & 0.036 & 0.000 & 0.013 & 0.000 \\
\bottomrule
\end{tabularx}\par}
\caption{Coherence error from noise alone, two main models (median [95\% range] over 1,000 simulations). For averaged answers no simulation reached the observed value.}
\label{tab:noise}
\end{table*}

No simulated coherent model with the models' run-to-run noise reached the observed coherence error of their averaged answers (Table~\ref{tab:noise}). To measure how large a coherence error run-to-run noise alone would produce, we made each model's averaged answers add up exactly, so that the model is perfectly coherent, added noise back and measured the coherence error, 1,000 times. We used three noise models. \emph{Resampling} draws each area's own log deviations from its $r$-run mean, scaled by $\sqrt{r/(r-1)}$ because deviations from a mean of $r$ runs understate single-run spread; Tables~\ref{tab:s1settings} and~\ref{tab:s2} use this one. \emph{Own variance} adds normal noise on the log scale with each area's own run-to-run variance. \emph{One variance} uses one variance per geography level, matched to the median run-to-run difference; it is the most conservative of the three for averaged answers. For an average of $r$ runs the draws are averaged (resampling) or the variance is divided by $r$. Table~\ref{tab:noise} gives the median [95\% range] for the two main models of Study~1, and the share of simulations at or above the observed value, the largest over the three noise models. For single runs the share is at most 0.036 (gpt-6.1-sol) and 0.013 (grok-4.7). The single runs gave coherence errors of 12\%, 13\% and 19\% (gpt-6.1-sol) and 17\%, 16\% and 17\% (grok-4.7).

\section{Recall Baseline}\label{app:recall}

Averaged answers in Study~1 sit closer to the 2020 count than to the ACS figure in most areas. To test whether that points to recall, we compared them with an estimator that recalls neither figure but is as accurate as the model. For each area we drew an estimate whose log error relative to the 2020 count is normal with mean zero, independent of the gap between the ACS figure and the count, and scaled so that its median error against ACS matches the model's averaged answers (tract 15.0\% and block group 28.4\% for gpt-6.1-sol; 16.7\% and 32.2\% for grok-4.7). The baseline is the share of draws closer to the count than to the ACS figure, computed over all Cook County tracts and block groups and weighted to our mix of 30 tracts and 89 block groups: 60\% for gpt-6.1-sol and 58\% for grok-4.7 (58--60\% across the nine settings). The observed shares, 60\% and 61\% (52--63\% across settings), are within three points of these baselines for the main models, as accuracy alone predicts.

\section{Study 2: Pre-specified Tests}\label{app:s2}

\begin{table*}[tp]
\centering
{\footnotesize
\begin{tabularx}{\textwidth}{@{}p{4.6cm}CCCp{2.3cm}@{}}
\toprule
Test & gpt-6.1-sol, low & gpt-6-luna, low & gpt-5.4, none & Result \\
\midrule
Coherence error above the 97.5th percentile of noise alone & 16.0 / 16.2\% vs 6.0 / 6.0\% & 17.7 / 17.6\% vs 8.3 / 8.6\% & 27.2 / 28.6\% vs 11.2 / 11.2\% & Supported, 6 of 6 \\
Coherence error increases with $k$: slope of $\ln$(sum / tract answer) on $\ln k$ above 0 [95\% CI] & 0.29 [0.20, 0.36] / 0.35 [0.29, 0.42] & 0.31 [0.27, 0.35] / 0.31 [0.28, 0.35] & 0.36 [0.27, 0.43] / 0.37 [0.28, 0.45] & Supported, 6 of 6 \\
Same area, lower as a block group: block-group answer below the tract answer in single-block-group tracts (median change; $n$) & $-$17 / $-$17\%; 58 / 57 & $-$27 / $-$27\%; 58 / 58 & $-$29 / $-$29\%; 60 / 60 & Supported, 6 of 6 (Holm $p < 10^{-7}$) \\
Joint answers add up (median coherence error below 1\%) & 0.0 / 0.0\% & 0.0 / 0.0\% & 0.0 / 0.0\% & Supported, 6 of 6 \\
Joint answers rank block groups worse (median within-tract Spearman, as asked vs joint; 179 tracts with 3 or more) & 0.5 vs 0.5 & 0.5 vs 0.5; Holm $p = 1$ & 0.4 vs 0.5, better; Holm $p = 0.002$ & Not supported \\
Share of near-even joint splits & 4.5 / 6.0\% & 7.7 / 8.4\% & 8.1 / 7.9\% & -- \\
A correction fixed in advance beats the best simple method at both levels (held-out tracts) & \multicolumn{3}{c}{--} & Not supported (1 of 12 tests passed) \\
\bottomrule
\end{tabularx}\par}
\caption{Study 2: pre-specified tests, population / housing units. The correction test uses the held-out tracts, the others every sample tract with complete answers; the ranking test applies to gpt-6-luna and gpt-5.4.}
\label{tab:s2tests}
\end{table*}

The analysis plan fixed the hypotheses, tests and exclusion rules before any model was queried, and the correction method was fixed before its evaluation on the held-out tracts. Zero answers count as missing, as in Study~1; fewer than 1\% of answers were missing. Table~\ref{tab:s2tests} reports the pre-specified tests.

\end{document}